\documentclass[runningheads,oribibl]{llncs}

\usepackage{url}
\usepackage{graphicx}
\usepackage[cp1250]{inputenc}
\usepackage{url}
\usepackage[ruled,vlined]{algorithm2e}
\usepackage{array}
\usepackage{amsmath}
\newcolumntype{x}[1]{>{\raggedleft\hspace{0pt}}p{#1}}

\newenvironment{keywords}{
       \list{}{\advance\topsep by0.35cm\relax\small
       \leftmargin=1cm
       \labelwidth=0.35cm
       \listparindent=0.35cm
       \itemindent\listparindent
       \rightmargin\leftmargin}\item[\hskip\labelsep
                                     \bfseries Keywords:]}
     {\endlist}

\begin{document}

\bibliographystyle{ieeetr}

\title{Random Graph Generator for Bipartite\\ Networks Modeling}
\author{Szymon Chojnacki and Mieczysław Kłopotek}

\institute{Institute of Computer Science, \\Polish Academy of Sciences \\ J.K. Ordona 21, 01-237 Warsaw, Poland\\
 \email{\{sch, klopotek\}@ipipan.waw.pl}}

\maketitle

\begin{abstract}
The purpose of this article is to introduce a new iterative algorithm with properties resembling real life bipartite graphs. The algorithm enables us to generate wide range of random bigraphs, which features are determined by a set of parameters. We adapt the advances of last decade in unipartite complex networks modeling to the bigraph setting. This data structure can be observed in several situations. However, only a few datasets are freely available to test the algorithms (e.g. community detection, influential nodes identification, information retrieval) which operate on such data. Therefore, artificial datasets are needed to enhance development and testing of the algorithms. We are particularly interested in applying the generator to the analysis of recommender systems. Therefore, we focus on two characteristics that, besides simple statistics, are in our opinion responsible for the performance of neighborhood based collaborative filtering algorithms. The features are node degree distribution and local clustering coefficient. 


%

\end{abstract}

\section{Introduction}
The analysis of large networks is driven by the desire to understand and model as diverse phenomena as the spread of infection, social communities creation, protein interactions or website importance assessment \cite{Newman_2010}. The interest of research community in complex networks was fueled by an empirical evidence which proved that some properties of real-life graphs are unachievable for classic random models. Moreover, the similar properties are common to networks observed in various fields. Several statistics describing networks can be measured. However, node degree distribution and mean clustering coefficient are two measures of a great importance. They are correlated for example with such macro features as an average length of a path between two nodes, the network's resilience to an attach or the pace of spread of innovations. It turns out that in diverse real-life networks:
\begin{itemize}
	\item node degree distribution is heavy-tailed
	\item mean clustering coefficient is bounded away from zero
\end{itemize}

In the classic theory of random graphs developed by two Hungarian mathematicians Paul Erd\H{o}s and Alfr\'{e}d R\'{e}nyi \cite{Erdos_1960} the asymptotic node degree distribution is \textit{Poisson}. Also the value of clustering coefficient, which measures the probability that two nodes sharing a friend are connected differs from empirical results and tends to zero as a number of nodes grows.

The seminal paper of Barab\'{a}si and Albert \cite{bar99a} describes the driving forces which are responsible for the heavy-tailed node degree distributions. The property can be attributed to both: the growth and the preferential attachment mechanism. Moreover, none of the two results in the desired distribution on its own. Kumar and collaborators \cite{kum00} proposed to substitute the preferential attachment mechanism with random selection of a neighboring node, which also leads to the heavy-tailed distribution. Liu \cite{liu_2002} described how a mixture of preferential and random attachment enables us to generate networks with weakened heavy-tail. V\'azquez \cite{vazquez} proposed a random graph generative procedure which results in networks with positive values of the clustering coefficient. The combined translation of the four results onto the ground of bigraphs comprises the frame of our algorithm.

Recently a few random bipartite graph generating algorithms have been introduced (\cite{Zheleva_2009},\cite{guillaume_2004}, \cite{lattanzi_2009}, \cite{chojnacki_2010}). However, none of them enables to generate growing networks with varying distributions and clustering coefficient bounded away from zero.
 
Our contribution comprises four main results:
\begin{enumerate}
	\item definition and formal justification of new local clustering coefficient dedicated for bigraphs - bipartite local clustering coefficient (BLCC)
	\item introduction of \textit{bouncing mechanism} responsible for the growth of BLCC
	\item description and analysis of new versatile bigraph generator
	\item identification of a relationship between network properties of bigraphs and the properties responsible for the complexity of recommender systems 
\end{enumerate}

The rest of the article is organized as follows. In Section 2 we formalize node degree distributions, local clustering coefficient and introduce BLCC. In Section 3 we outline the motivation for our research, which is based on the equivalence of bipartite graphs and user-item matrices in the recommender systems. The fourth section contains a description of our algorithm. In Section 5 we present the results of numerical simulations. The last sixth section is dedicated for the concluding remarks. Advanced mathematical transformations are described in details in two appendices.
\newpage

\section{Background}

A graph is an ordered pair $G=(V,E)$ comprising a set of vertices $V$ and a set of edges $E\subseteq\{V×V\}$. A bipartite network  is a graph $G=(U\cup I,E)$ which vertices can be labeled by two types $U$ and $I$. The difference with a classic unipartite graph is the fact that $V$ consists of two disjoint sets $V=\{U \cup I,U \cap I=\emptyset\}$ and edges exist only between nodes of different types $E\subseteq\{U×I\}$. We analyze undirected graphs.

\subsection{Node degree}

\begin{figure}[htbp]
	\centering
		\includegraphics[width=0.75\textwidth]{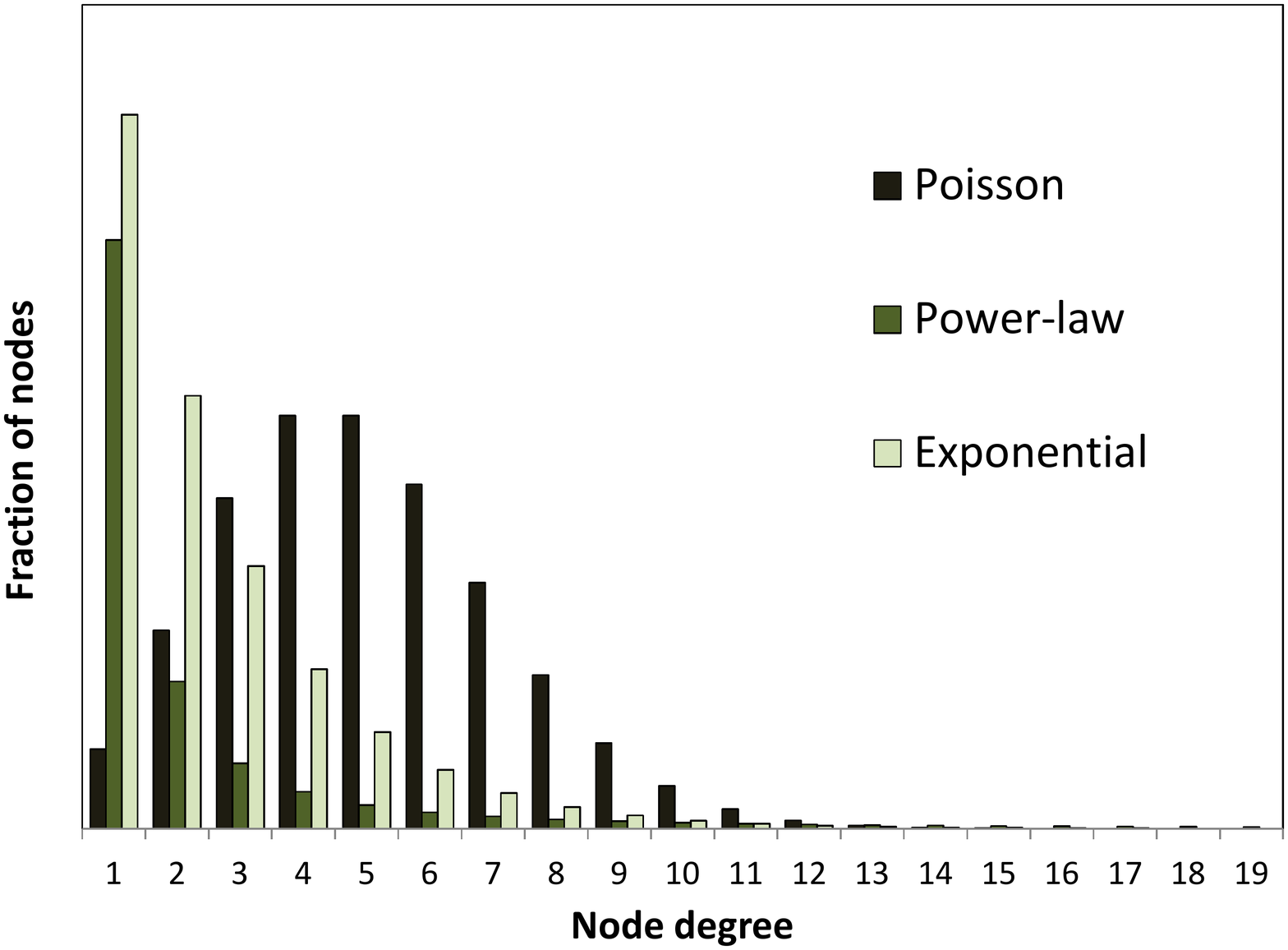}
	\caption{Three degree distributions with the same average. The Poisson distribution is characteristic for classic random graphs. The exponential and the power-law distributions are more common in real datasets. Both of them are skewed. However, the tail of the power-law distribution decays slower.}
	\label{fig:3distr}
\end{figure}

\begin{figure}
\begin{tabular}{cc}
\centerline{\includegraphics[width=0.80\textwidth]{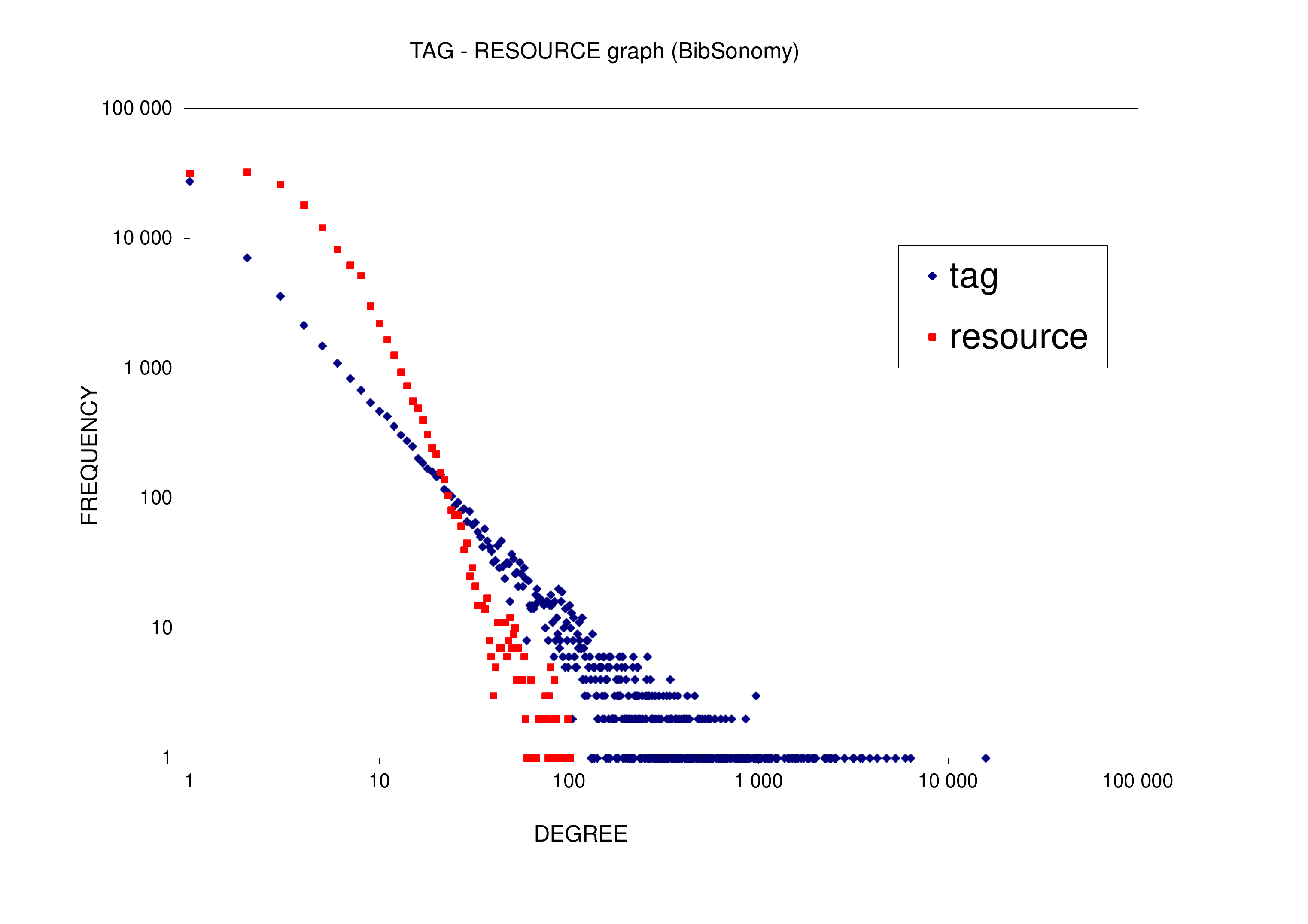}} \\
\centerline{\includegraphics[width=0.80\textwidth]{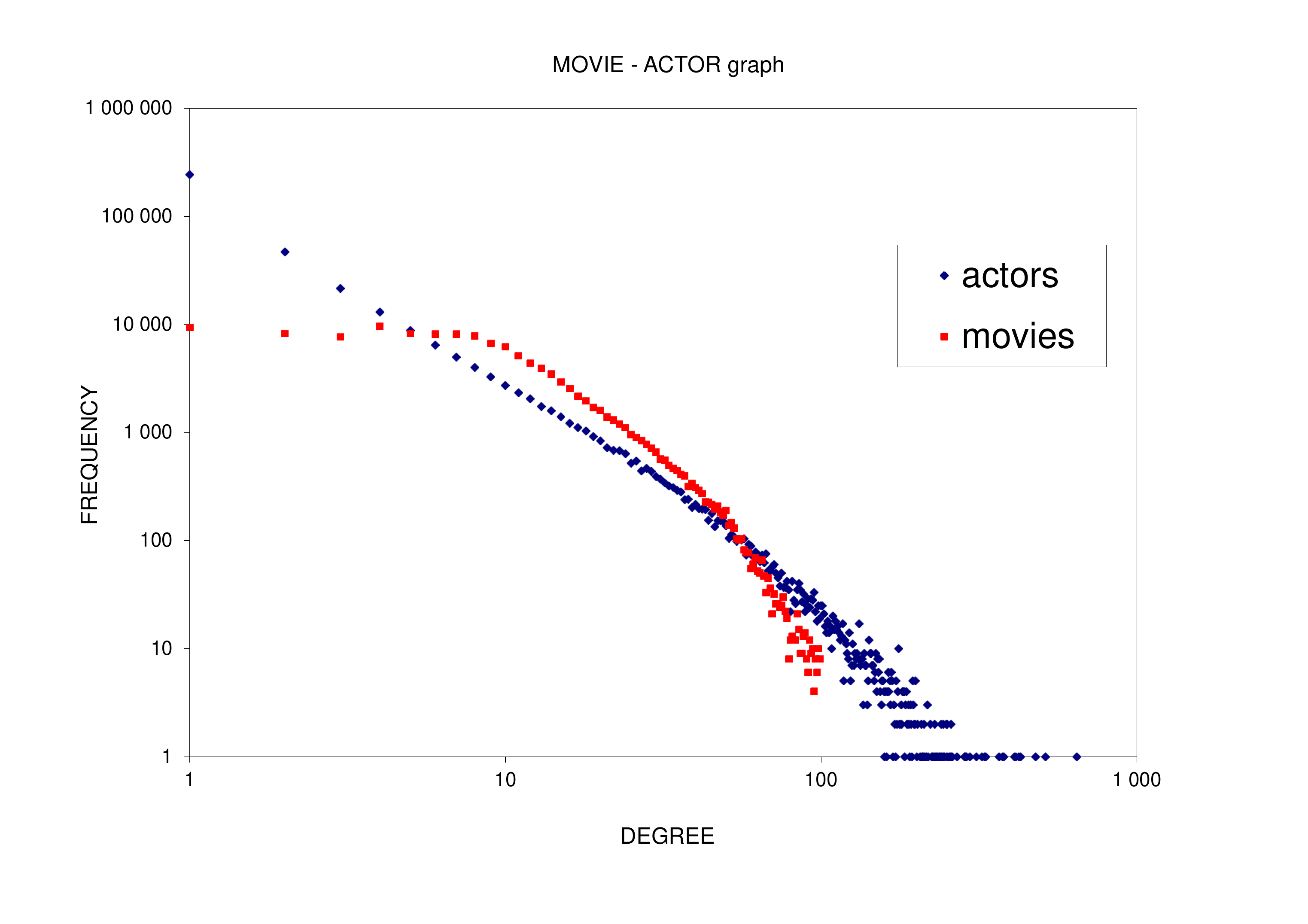}} \\
\centerline{\includegraphics[width=0.80\textwidth]{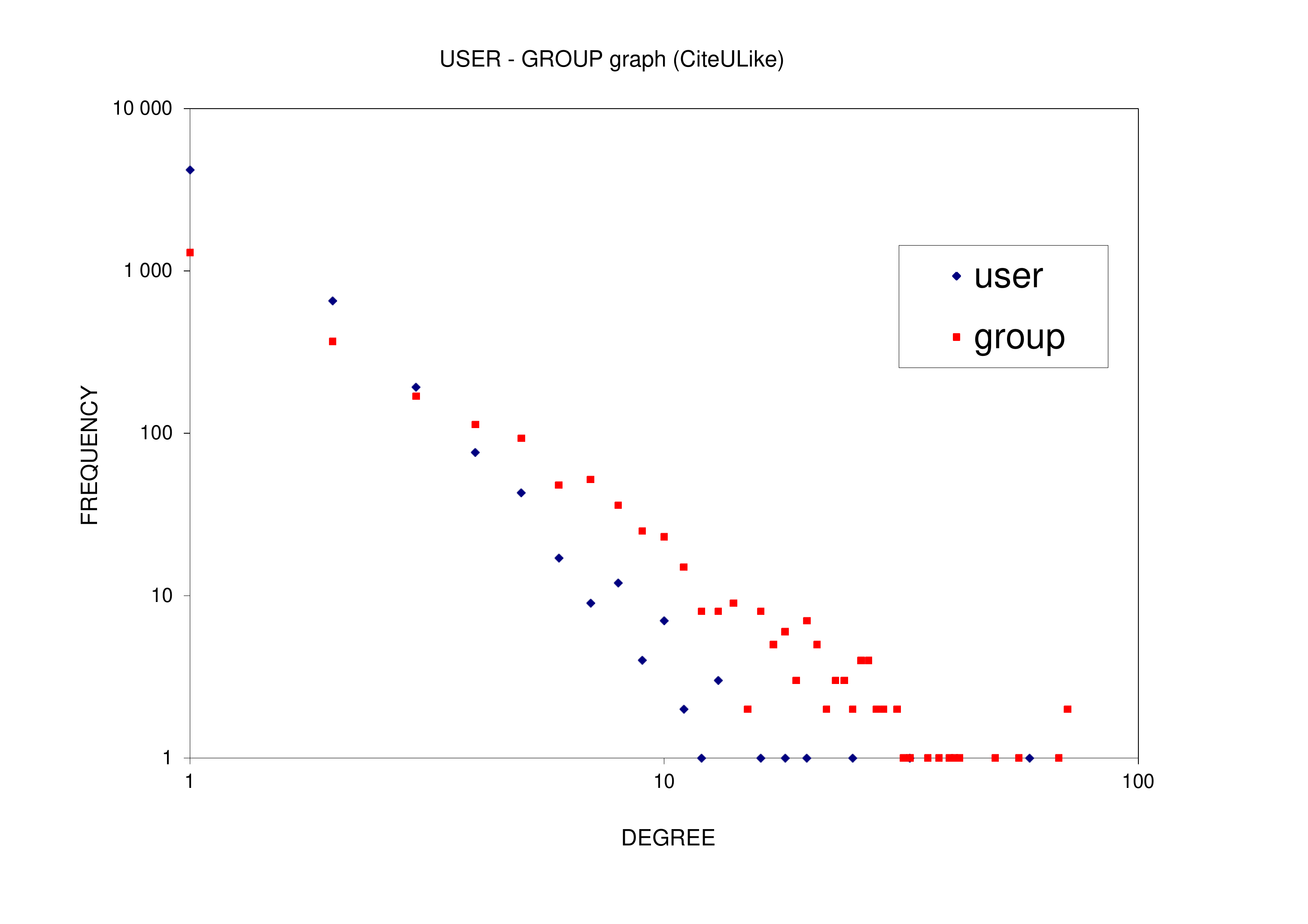}} 
\end{tabular}

\caption{The node degree distributions of three bipartite graphs. The straight line of points (on a LOG-LOG scale) in all three datasets envisions the power-law feature of the datasets. In case of BibSonomy \cite{ecml} (upper chart) and IMDB \cite{imdb} (middle chart) graphs, one modality tends towards exponential distribution. In case of CiteULike \cite{citeulike} (lower chart) dataset both modalities are shaped in a straight line.}
	\label{fig:distr}
\end{figure}

A degree of a node stands for the number of direct (first) neighbors of the node and is equal to the number of node's edges. The probability density function (pdf) of node degree distributions in real-life datasets is usually skewed (Fig. \ref{fig:distr}). If the tail decays slowly we can observe the power-law distribution $pdf_{PL}(x)=a  x^{-k}$. The tail vanishes quickly in the exponential distribution $pdf_{EX}(x)=\lambda  e^{-\lambda x}$. It is convenient to visualize the two distributions on a log-log scale. From the fact that $\log(pdf_{PL}(x))=-k  \log(x)+log(a)$ follows that the power-law distribution is shaped in a straight line on a log-log chart. This distribution is called \textit{scale-free} because $pdf_{PL}(cx)=a  (cx)^{-k}=a c^{-k}pdf_{PL}(x)$. The distributions observed in real networks can not be generated by classic random graphs. The graphs studied by Erd\H{o}s give the Poisson distribution. The three types of distributions are drawn in Fig. \ref{fig:3distr}.

\subsection{Local clustering coefficient}
Local clustering coefficient is used to measure the probability that if two nodes share a neighbor than they are also connected. It is computed for each node and an average over all nodes indicates the level of network's transitivity. Let's denote by $c_j$ the number of connected pairs among the direct neighbors of node j and by $k_j$ the degree of node $j$. The local clustering coefficient ($LCC$) is given by: 

\begin{equation}
LCC_j=\frac{ c_j}
{k_j(k_j-1)/2}.
\label{lcc}
\end{equation}

The value of LCC is zero for any node in a bipartite graph. Therefore, we introduce a new coefficient dedicated to measuring transitivity in bigraphs. Bipartite local clustering coefficient ($BLCC$) of node $j$ takes values of one minus the proportion of node's second neighbors to the potential number of the second neighbors of the node. The value of $BLCC$ calculated for node $j$ is given by: 

\begin{equation}
BLCC_j=1-\frac{ |N_2(j)|}
{\sum_{i \in N_1(j)}{(k_i - 1)}},
\label{blcc}
\end{equation}

\noindent where $|N_2(j)|$ stands for the number of the second neighbors of node $j$, $N_1(j)$ is a set of the first neighbors of node $j$.

\begin{figure}[htbp]
	\centering
		\includegraphics[width=\textwidth]{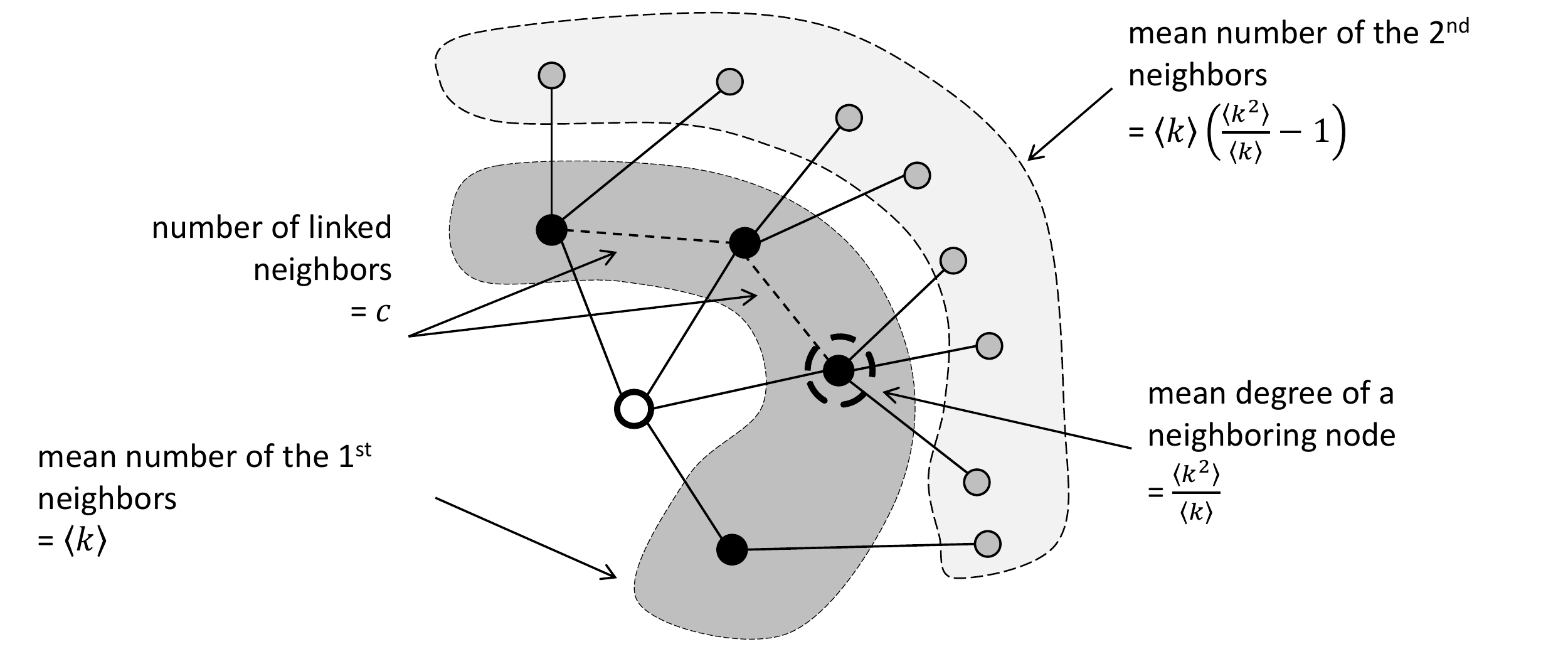}
	\caption{In order to compute the BLCC for a unipartite graph we need to assess the potential number of the second neighbors of a given node. A random node has $\langle k \rangle$ neighbors (in the figure $\langle k \rangle=4$). There are $c$ connections among the neighbors on average ($c=2$). Each neighbor has on average $\frac{\langle k^2 \rangle}{\langle k \rangle}$ edges. Each edge points to a second neighbor of the considered node or to the node ($\langle k \rangle$ edges) or to the first neighbor ($2c$ edges). We assume that there are no two different edges pointing to the same second neighbor.}
	\label{fig:clustering}
\end{figure}

In order to justify the correlation between $LCC$ and $BLCC$, we consider the values of the two coefficients in case of a unipartite graph. We denote by $f(c)$ in Eq. (\ref{fc}) the value of $LCC$ calculated for a random node with $c$ pairs of connected neighbors. We use $g(c)$ in Eq. (\ref{gc}) to assess the value of BLCC in case of the same node. Except of $c$ pairs we follow the tree like structure assumption. We substitute $k_i$ with $\frac{\langle k^2 \rangle}{\langle k \rangle}$ (i.e. the expected degree of a \textit{neighboring node\footnote{The formula for an average degree of a neighboring node is derivated in appendix A.}} \cite{vega_2007}) and observe that on average $|N_1(j)|=\langle k \rangle$. The logic of deriving $|N_2(j)|$ is presented in Fig. \ref{fig:clustering}. 

\begin{equation}
f(c)=\frac{2c}{\langle k \rangle \left( \langle k \rangle -1 \right)}=\frac{2c}{\langle k \rangle^2 - \langle k \rangle}
\label{fc}
\end{equation}

\begin{equation}
g(c)=1-\frac{\langle k \rangle \left( \frac{\langle k^2 \rangle}{\langle k \rangle} -1 \right) -2c}
{\langle k \rangle \left( \frac{\langle k^2 \rangle}{\langle k \rangle} -1 \right)}=\frac{2c}{\langle k^2 \rangle - \langle k \rangle}
\label{gc}
\end{equation}

From the fact that the variance of any distribution is nonnegative and it can be decomposed as $\sigma^2=\langle k^2 \rangle - \langle k \rangle^2$, we assert that $g(c)/f(c)$ is constant and not larger than one.

We also considered a different definition of the number of potential second neighbors in Eq. \ref{blcc}. Within the local tree-like structure setting \cite{Newman_2001} it can be approximated by $\langle u \rangle\left(\frac{\langle v^2 \rangle}{\langle v \rangle}-1\right)$. Even though on average such definition gives positive fractions (Table \ref{tab:tabelka1}), a value of BLCC calculated for one node can be negative and therefore we stay with the definition of BLCC as it is in Eq. \ref{blcc}.

\begin{table}[htbp]
	\centering
	\renewcommand{\arraystretch}{1.5}
{\scriptsize

		\begin{tabular}{l x{1.5cm} x{1.5cm} x{1.5cm} c x{1.5cm} x{1.5cm} x{1.5cm}}
				\hline
				 & \multicolumn{3}{c}{basic statistics} && \multicolumn{3}{c}{second neighbors} \tabularnewline
				\cline{2-4} \cline{6-8}
				& \multicolumn{1}{r}{users} & \multicolumn{1}{r}{items} & \multicolumn{1}{r}{edges} && \multicolumn{1}{r}{real} & \multicolumn{1}{r}{theoretic } & \multicolumn{1}{r}{$\frac{real}{theoretic}$} \tabularnewline
				\cline{1-4} \cline{6-8}
				
				CEO \cite{wasserman_faust94} & 26 & 15 & 98 && 21.8 & 22.0 & 0.99 \tabularnewline
				CiteULike \cite{citeulike} &5 208&2 336&7 196&&14.2&23.9&0.59\tabularnewline
				BibSonomy \cite{ecml} & 3 617 & 93 756 & 253 366 && 500.4 & 6 579.2 & 0.08 \tabularnewline
				YouTube \cite{mislove_2007} & 94 238 & 30 087 & 293 360 && 1 269.6 & 2 101.3 & 0.60 \tabularnewline
				IMDB \cite{imdb} & 383 640 & 127 823 & 1 470 404 && 78.4 & 211.4 & 0.37 \tabularnewline
				Flickr \cite{mislove_2007} & 395 979 & 103 631 & 8 545 307 && 1 217.4 & 52 704.9 & 0.02 \tabularnewline
				LiveJournal \cite{mislove_2007} & 3 201 203 & 7 489 073 & 112 307 385 && 785 194.2 & 1 521 273.4 & 0.52 \tabularnewline
				Orkut \cite{mislove_2007} & 2 783 196 & 8 730 857 & 327 037 487 && 334 863.6 & 2 294 114.8 & 0.15 \tabularnewline
				
				\hline
		\end{tabular}
		}
	\caption{An average number of the second neighbors in eight real-life datasets is smaller than approximated by the Newman's asymptotic formula (theoretic value). The most significant shrinking is observed in the Flickr dataset. The shrinking is observed in both relatively small and very large datasets.}
	\label{tab:tabelka1}

\end{table}

\section{Recommender systems}

Recommender systems are an important component of the Intelligent Web. The systems make information retrieval easier and push users from typing queries towards clicking at suggested links. We experience real-life recommender systems when browsing for books, movies or music. The engines are an essential part of such websites as \textit{Amazon}, \textit{MovieLens} or \textit{Last.fm}. The interest of research community in the systems was fueled by the Netflix movie recommendation competition \cite{netflix}. During the challenge the state-of-art systems in terms of accuracy were developed.

\begin{figure}[htbp]
	\centering
		\includegraphics[width=0.80\textwidth]{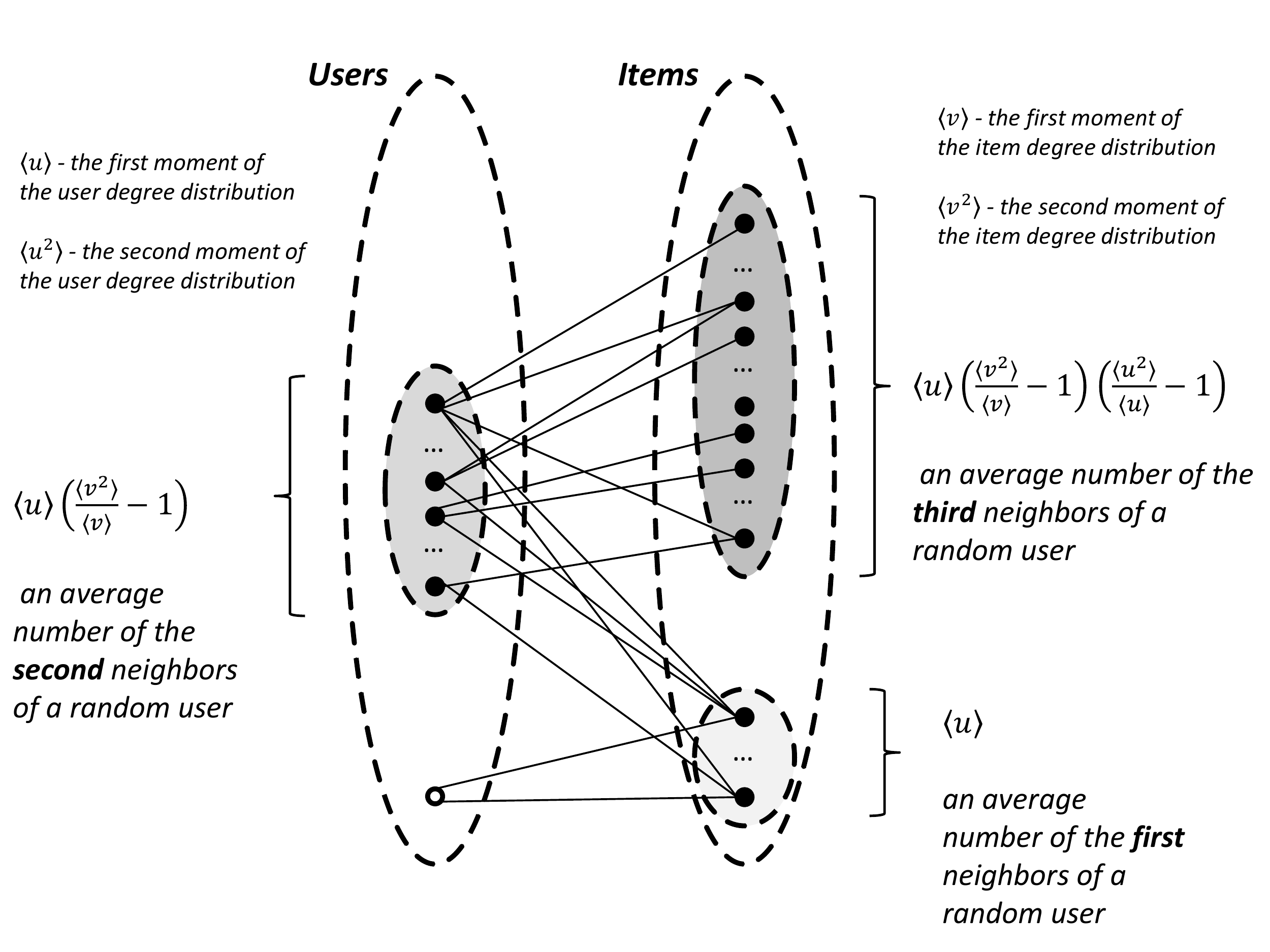}
	\caption{In recommender systems based on the neighborhood principle the recommended items are selected from the items of the users that have rated at least one common item with an analyzed user.}
	\label{fig:motivation}
\end{figure}

However, it has been shown recently during the ECML Discovery Challenge 2009 \cite{ecmlChallenge} that the most accurate recommender systems fail to meet real-life constraints. It is not an easy task to update trained models when new items or users enter the evaluation. The problem is usually referred to as the \textit{Cold Start} problem. These observations constitute the motivation for our research. We believe that there exists a need for algorithms that can generate random recommendation matrices (or equivalently bipartite graphs). We are particularly interested in the neighborhood-based techniques. These methods are the best suited for the dynamically changing scenarios, but the latency of creating a recommendation depends significantly on the structure of underlying dataset (compare Fig. \ref{fig:motivation}). Moreover, because of embedding iterative mechanism in our generator, it can be used to simulate the \textit{Cold Start} cases.

\section{Our algorithm}
Our algorithm consists of three steps: (1) new node creation, (2) edge attachment type selection and (3) running bouncing mechanism. The procedure requires specifying eight parameters:
\newline
\begin{tabular}{lcl}
$m$ & - & the number of initial loose edges with a user and an item at the ends\\
$T$ & - & the number of iterations\\
$p$ & - & the probability that a new node is a user\\
    &  & $(1-p)$ is the probability that a new node is an item\\
$u$  & - & the number of edges created by each new user\\
$v$  & - & the number of edges created by each new item\\
$\alpha$  & - & the probability that a new user's edge is being connected to\\
      &  & an item with preferential attachment\\
$\beta$  & - & the probability that a new item's edge is being connected to\\
      &  & a user with preferential attachment\\
$b$ & - & the fraction of preferentially attached edges \\
&&that where created via a \textit{bouncing mechanism} \\

\end{tabular}
\newline
Steps (1) and (2) are explained in Sec. 4.1 and analyzed in Sec. 4.2. In Sec. 4.3. step (3) is discussed.
 \subsection{Basic model}
In the basic model we utilize first seven parameters. The bouncing mechanism is applied in the full model as an additional third step.
\begin{figure}[htbp]
	\centering
		\includegraphics[width=0.65\textwidth]{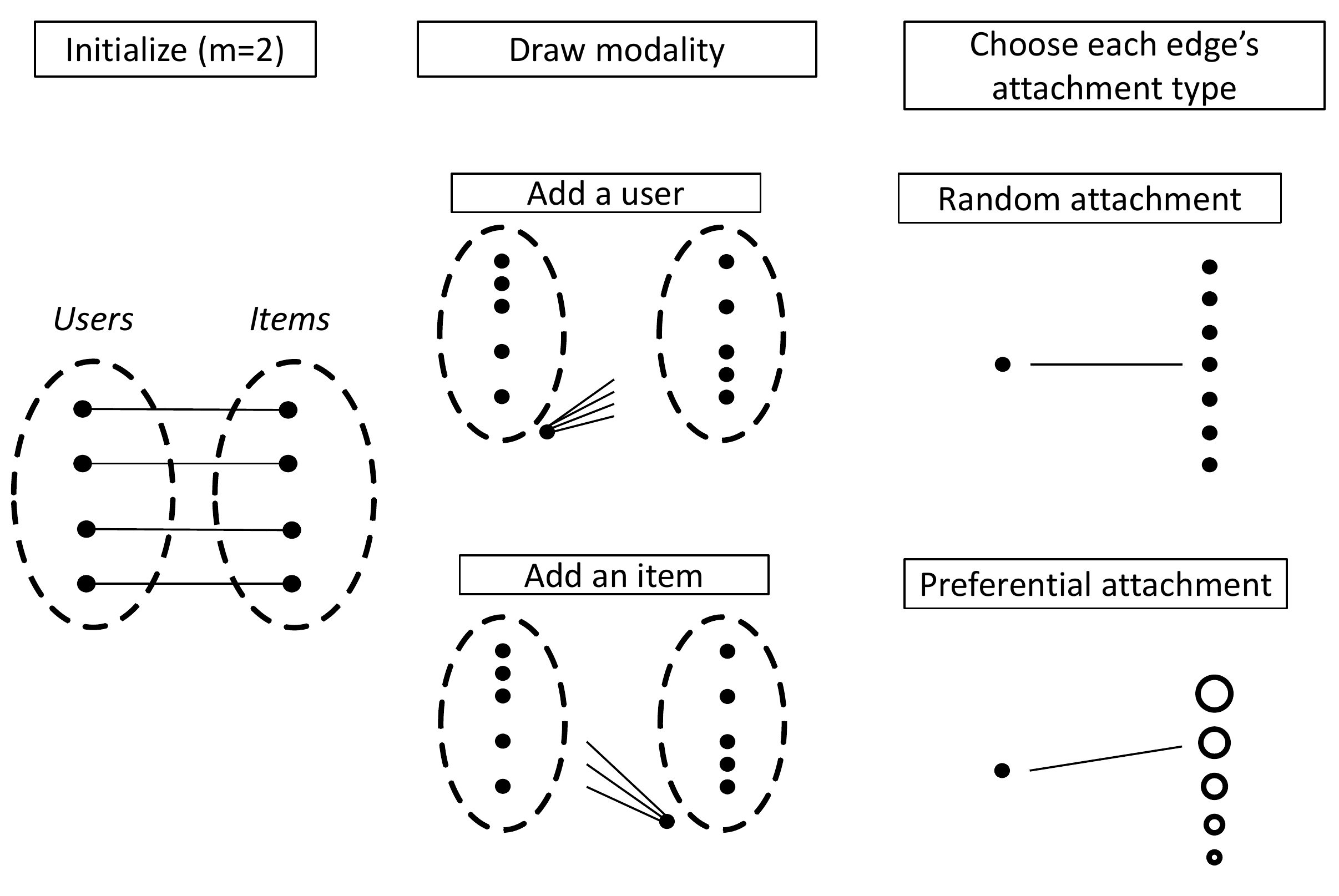}
	\caption{The bipartite random graph generator is initialized with a set of $m$ pairs of users and items. During each iteration two steps are performed. In the first step the type of new node is determined. In the second step a decision is made on the level of each node's edge whether to draw its ending with preferential attachment or randomly. In the preferential attachment variant the probability that a node is drawn is proportional to its degree.}
	\label{fig:stepsModel}
\end{figure}
\newline
The basic model is based on an iterative repetition of two steps (Fig. \ref{fig:stepsModel}).
\newline
\textbf{Step 1} If a random number is greater then $p$ create a new user with $u$ loose edges, otherwise create a new item with $v$ loose edges.
\newline
\textbf{Step 2} For each edge decide whether to join it to a node of the second modality randomly or with preferential attachment. The probability of selection preferential attachment is $\alpha$ for new user and $\beta$ for new item.

\subsection{Formal analysis}

One can see that after $t$ iterations the bigraph consists of $|U(t)|=2m+p  t$ users, $|I(t)|=2m+(1-p)  t$ items, and $|E(t)|=4m+t  (p  u+(1-p)  v)$ edges. Let's denote by $\eta$ an average number of edges created during one iteration $\eta=(p  u+(1-p)  v)$. After relatively many iterations $(t>>m)$ we can neglect $m$. In the presented model, an average user degree is:

$$\frac{|E(t)|}{|U(t)|}=\frac{4m+t  (p  u+(1-p)  v)}{2m+p  t}= \frac{\eta}{p},$$

\noindent analogously an average item degrees is:

$$\frac{|E(t)|}{|I(t)|}= \frac{\eta}{(1-p)},$$

\noindent the values are time invariant, but depend on both $u$ and $v$.

In the following deduction we look from user modality perspective. However, the computations can be altered to the opposite item modality easily. In order to derive asymptotic node degree distribution in our model we need to specify the probability that a user node $j$ with degree $k_j$ gets connected to a new item. The quantity is usually represented as $\Pi(k_j)$ within the complex networks community. If nodes are selected randomly than: 

$$\Pi_{random}(k_j)= \frac{1}{|U(t)|}=\frac{1}{pt}.$$

In case of random attachment $\Pi(k_j)$ does not depend on $k_j$. If nodes are selected with accordance to the preferential attachment rule than:

$$\Pi_{preferential}(k_j)=\frac{k_j}{|E(t)|}=\frac{k_j}{\eta t}.$$
 
Contrary to the random attachment scenario, the probability of node's selection is linearly proportional to its current degree. The probability of drawing a node with degree $k_j$ is the degree divided by the number of edges. We can verify that by summing the values of $\Pi$ over all user nodes we get one $\sum_j{\Pi_j}=1$. In our model the decision whether to draw a user for an item with random or preferential attachment depends on $\beta$, hence the combined formula is:
 \begin{equation}
\Pi(k_i)=\beta \frac{1}{p  t} + (1-\beta)\frac{k_i}{\eta t}.
\label{basicPI}
\end{equation}
 
 The equation (\ref{basicPI}) enables us to describe the pace of growth of nodes all with degree $k_i$ as

\begin{equation}
\frac{\partial k_i}{\partial t}=(1-p) v  \Pi(k_i).
\label{basicDif}
\end{equation}

We assume in the above equation that time interval between iterations is very small and that all nodes with a given degree grow in the same way. We show in the appendix that

\begin{equation}
P(k)\propto \left( \frac{\beta \eta +p(1-\beta)k}{\beta \eta + p(1-\beta)u} \right)^{\frac{-\eta}{(1-p)(1-\beta)v}-1}.
\label{distr4}
\end{equation}

One can verify that for $\beta=0$ we get power-law distribution. If $\beta \rightarrow 1$, we can utilize the fact that $\lim_{n\rightarrow \infty}\left( 1+\frac{c}{n}\right)^n=e^c$ in order to obtain exponential distribution. The above result is consistent with \cite{bar99a}. When we put $\beta=0$, $p=0.5$ and $u=v$ we have power-law distribution with the scaling exponent equal to $3$.

\subsection{Full model}

\begin{figure}[htbp]
	\centering
		\includegraphics[width=0.9\textwidth]{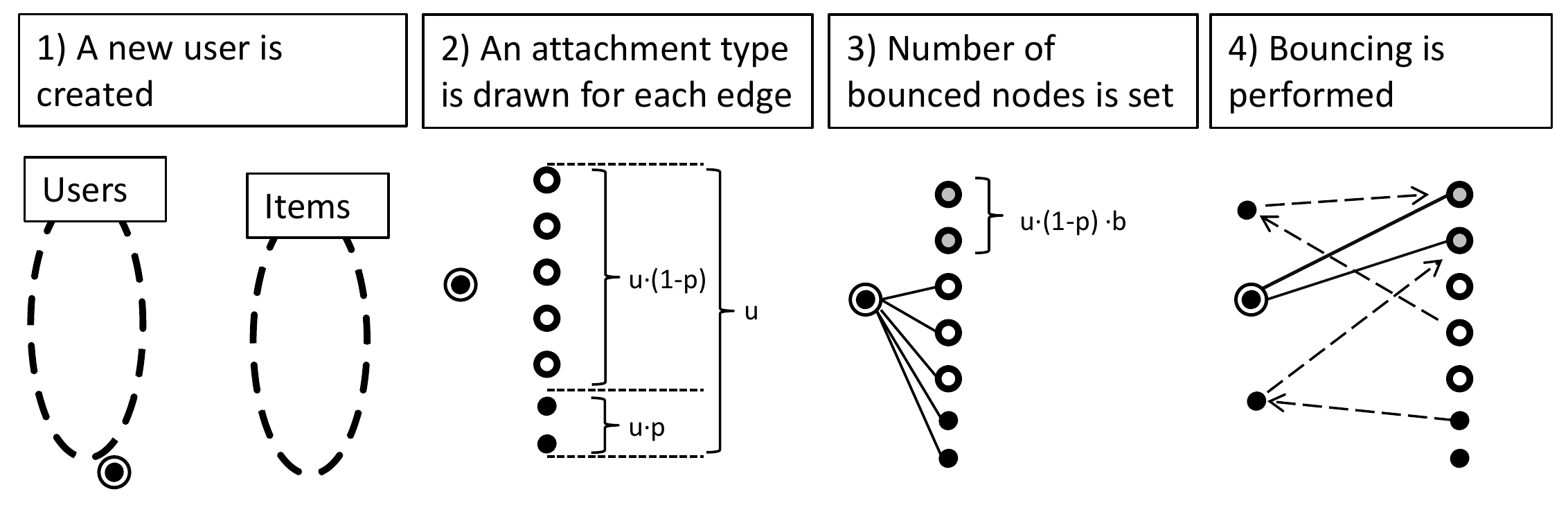}
	\caption{For each edge of a new node, that is to be connected with an existing node with accordance to the preferential attachment mechanism, a decision is made whether to create it via a bouncing mechanism. In case of attaching new user node, $u$ new edges are created. On average $u\cdot\alpha$ edges' endings are to be drawn preferentially and $u\cdot\alpha\cdot b$ of them are to be obtained via bouncing from the nodes that are already selected. }
	\label{fig:bouncing}
\end{figure}

We have shown recently that node degree distributions of both modalities can be responsible for BLCC in some networks, but in others there exist additional shrinking forces responsible for high values of BLCC \cite{chojnacki_2010a}. Therefore we introduce the \textit{bouncing mechanism} (Fig. \ref{fig:bouncing}), which is based on surfing the web technique \cite{vazquez}. The mechanism enables us to rise BLCC, but can only by applied to the edges that are to be selected with preferential attachment. This can be attributed to the fact that the probability that a random walk is finished in a node is proportional to its degree \cite{burda_2009}. Bouncing is performed in three micro steps: (1) a random node is drawn from the nodes that are already joined with the new node, (2) a random neighbor of the drawn node is chosen, (3) a random neighbor of the neighbor is selected for joining with the new node. 

\begin{table}
	\centering
		\begin{tabular}{c}
			\\
			\\
			\\
			\\
			\\
			\\
			\\
			
		\end{tabular}
\end{table}

\IncMargin{1em}
\begin{algorithm}

\SetKwData{Users}{Users}
\SetKwData{Items}{Items}
\SetKwData{NewUser}{NewUser}
\SetKwData{NewItem}{SelectedItem}

\SetKwData{TempItems}{TempItems}
\SetKwData{Edges}{Edges}

\SetKwData{SelectedGroups}{SelectedGroups}

\SetKwFunction{DrawItemPreferentially}{DrawItemPreferentially()}
\SetKwFunction{DrawItemRandomly}{DrawItemRandomly}
\SetKwFunction{DrawTempNode}{DrawTempNode}
\SetKwFunction{BounceFromRandom}{BounceFromRandom}
\SetKwFunction{RAND}{RAND()}
	\eIf{\RAND $\leq p$ }{
	\tcp{$p$ - the probability that a new node is a user}
	\BlankLine
		\For{$k \leftarrow 1$ \KwTo $u$}{
		\tcp{$u$ - the number of edges created by anew user}
		\BlankLine
			\eIf{\RAND $\leq \alpha$}{
			\tcp{$\alpha$ - the probability that the new user's item is drawn preferentially}
			\BlankLine
				\eIf{\RAND $\leq b$}{
					\tcp{$b$ - the probability that new preferential node was chosen by bouncing}
					\BlankLine
					\NewItem $\leftarrow$ \BounceFromRandom{\TempItems} \;
				}
				{
				\NewItem $\leftarrow \DrawItemPreferentially$ \;
				\TempItems $\leftarrow \NewItem$ \;	
			  }
			}{
				\NewItem $\leftarrow \DrawItemRandomly$ \;
				\TempItems $\leftarrow$ \NewItem \;	
			}
		}
	\Users $\leftarrow$ \Users $\cup$ \NewUser \;	
	\Edges $\leftarrow$ \Edges $\cup \{ \TempItems \times \NewUser \}$ \;
	}{
			\emph{Process analogously with new item node}
	}

\caption{An iteration of the bipartite graph generator}
\label{algo}
\end{algorithm}
\newpage

\section{Numerical results}

The results of the numerical experiments are divided into three subsections. In the first part we shortly present a Java applet developed in our Lab to play with various parameters of the generator. In the second part we show which parameters impinge on the values of node degree distributions and BLCC. In the last section we show how the number of potentially similar users and the number of their items can be determined by various levels of the generators parameters.
\subsection{Graphical analysis}
\begin{figure}[htbp]
	\centering
		\includegraphics[width=0.95\textwidth]{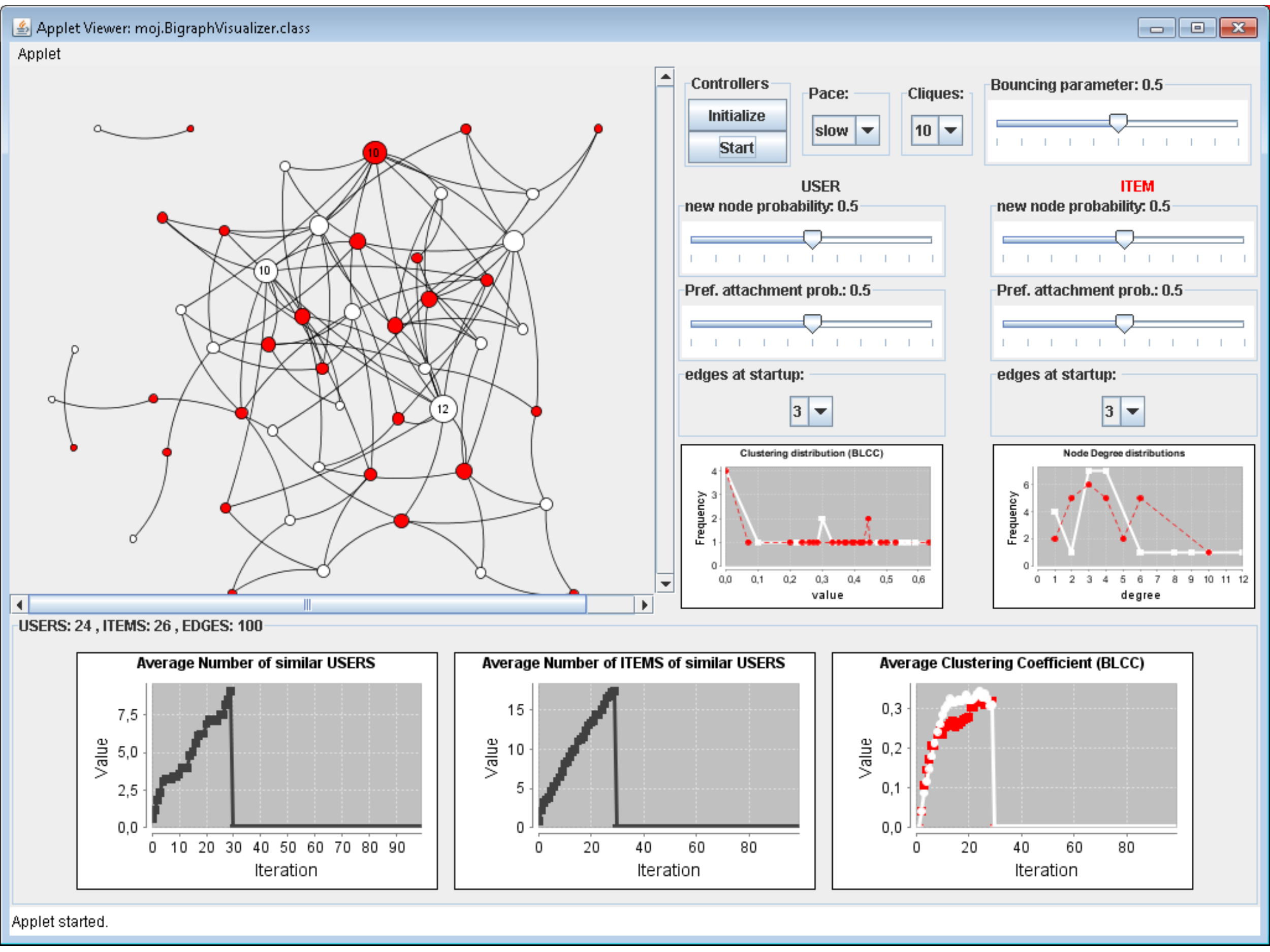}
	\caption{A bigraph generated after $t=30$ iterations. The values of all probabilities were set to $0.5$, each new node creates three new edges $u=v=3$, initial number of pairs $m=10$.}
	\label{fig:screen3}
\end{figure}

The applet presented in Figure  \ref{fig:screen3} can be accessed online in \url{http://www.ipipan.eu/~sch/software/applet.html}. All parameters (except of the initial number of pairs) can be changed during graph generation. The distributions of BLCC and node degrees are being updated online for both modalities. Alse the average number of potentially similar users and their items is visualized at a chart. By an expression \textit{similar user} we understand all users that have rated at least one item in common with the selected user.

\subsection{Social network properties}
\begin{figure}[htbp]
\begin{tabular}{lr}
	\centering
\includegraphics[width=0.5\textwidth]{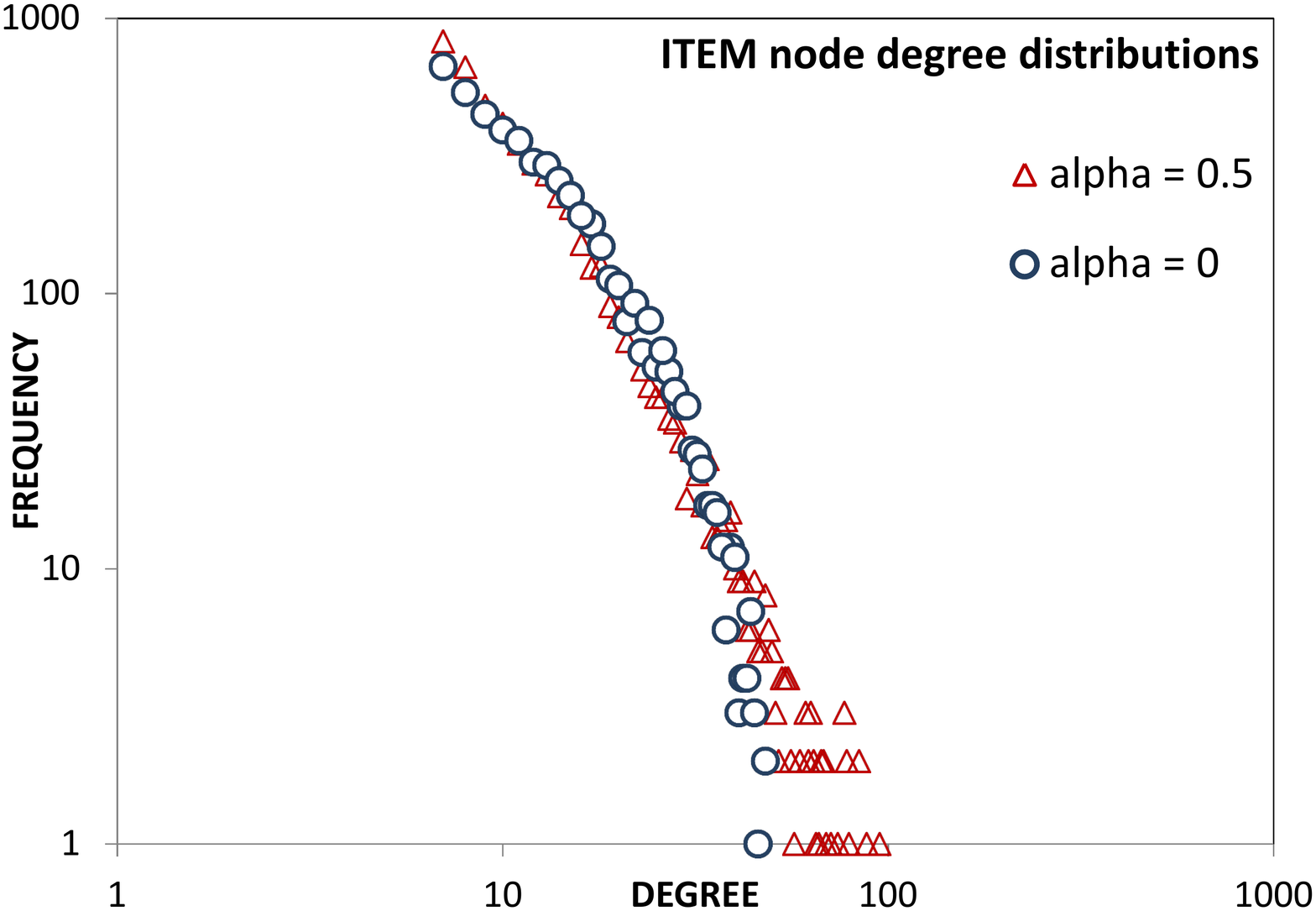}&\includegraphics[width=0.5\textwidth]{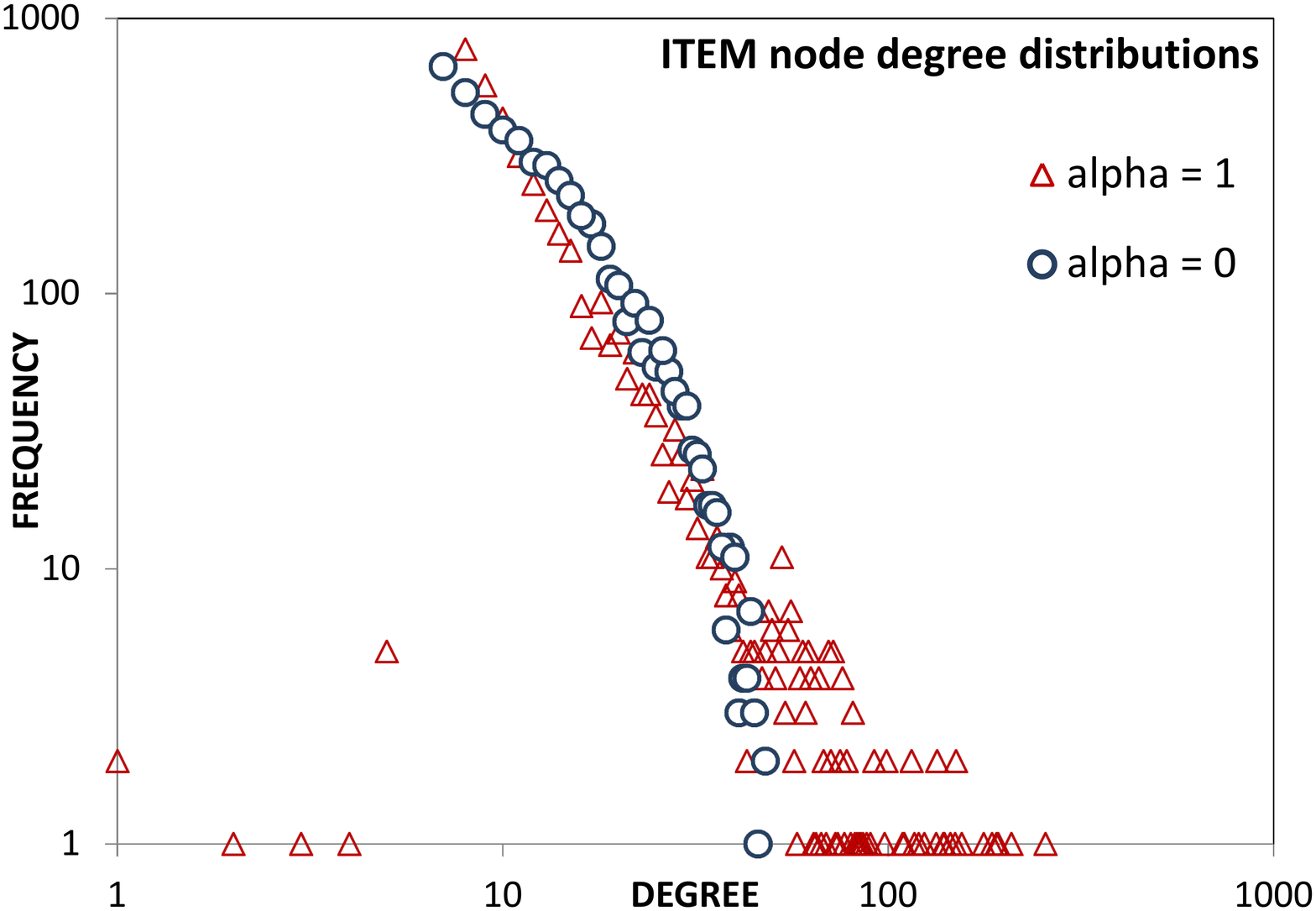}\\
	\end{tabular}
	\caption{Left Panel: blue circles indicate that the random attachment of users' edges (i.e. items) results in the exponential distribution of item degrees. Red triangles in both panels show that as $\alpha \rightarrow 1$ the distribution becomes power-law. Experiments run with ($m=50$, $T=10~000$, $p=0.5$, $u=v=7$, $\beta=0.5$).}

	\label{fig:distrUSER_ITEM}
\end{figure}

We consider node degree distributions of both modalities and the values of BLCC as the network properties of the generated graphs. Node degree distributions are controlled by two parameters: $\alpha$ and $\beta$. We show in Figure \ref{fig:distrUSER_ITEM} that if one parameter tends to one, the shape of appropriate modality becomes power-law. Low values output exponential distribution. Moreover, we do not observe any correlation between the distributions of both modalities.

\begin{figure}[htbp]
\begin{tabular}{cc}
	\centering
	\includegraphics[width=0.5\textwidth]{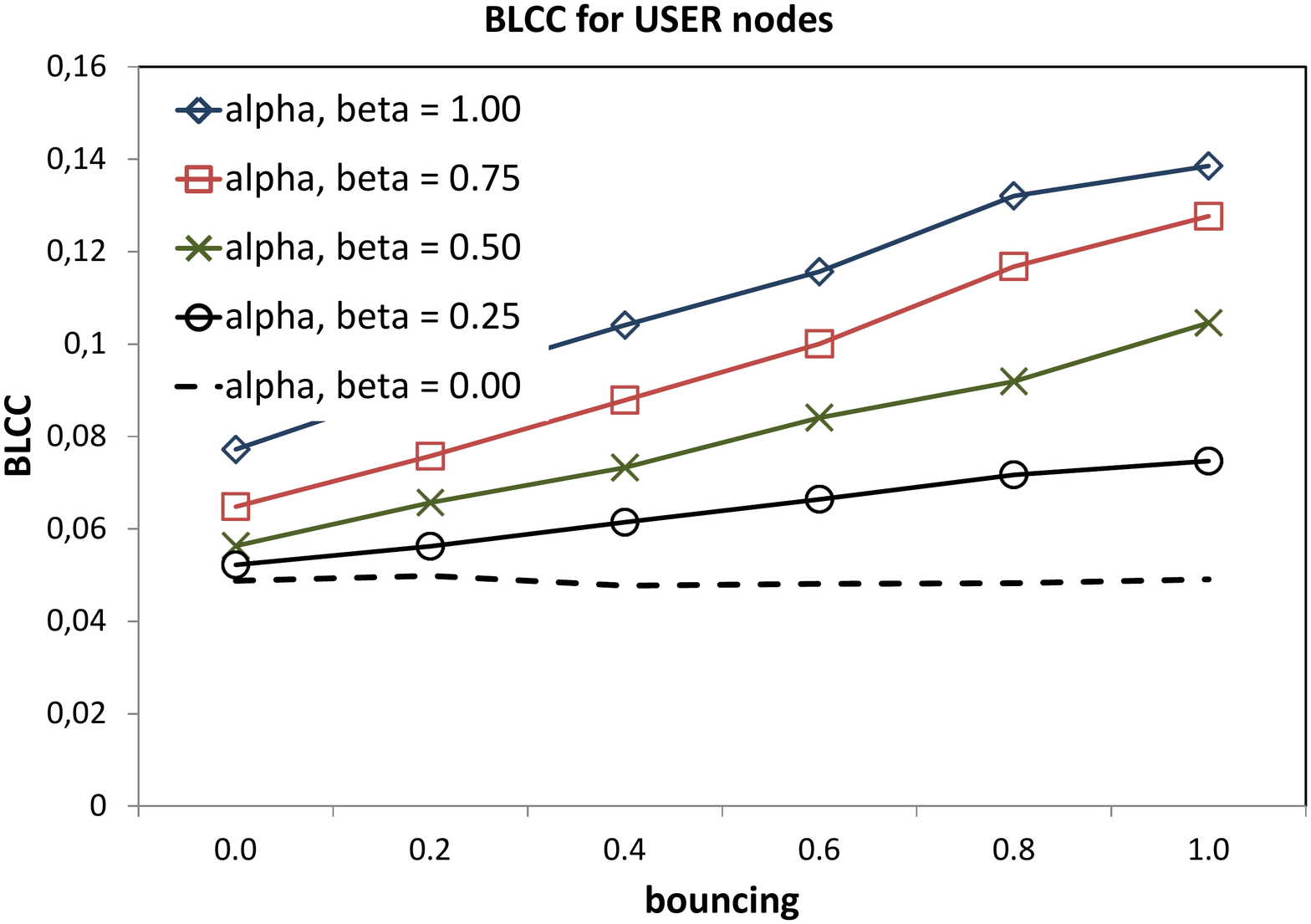}&\includegraphics[width=0.5\textwidth]{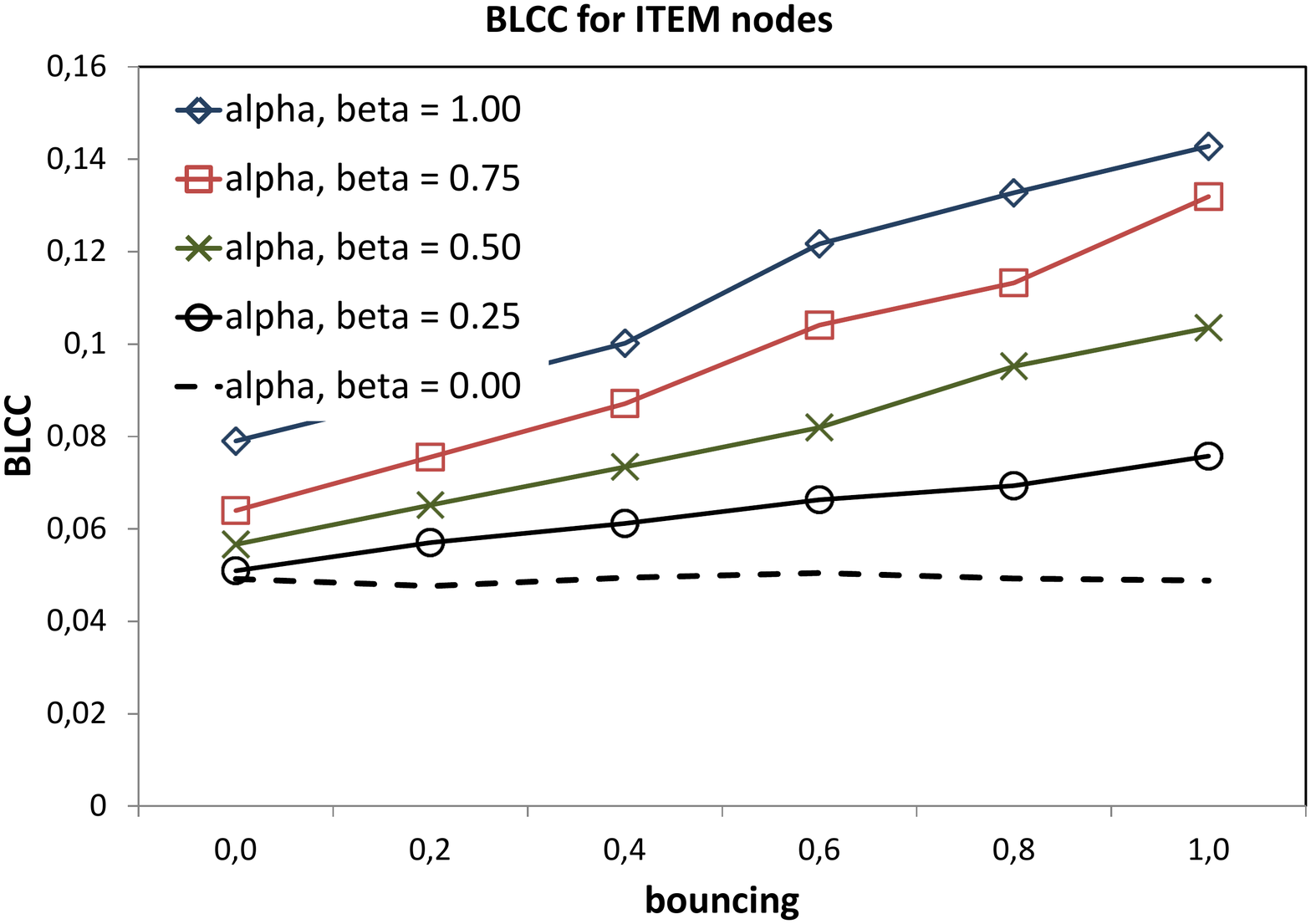}\\
	\end{tabular}
	\caption{The growth of the bouncing parameter $b$ results in higher values of BLCC (bipartite local clustering coefficient). If no nodes are connected with accordance to the preferential attachment mechanism $\alpha=\beta=0$, the values of $b$ do not influence BLCC. Experiments run with ($m=50$, $T=10~000$, $p=0.5$, $u=v=7$).}
	\label{fig:BLCC_by_bounce}
\end{figure}

The values of BLCC (\textit{bipartite local clustering coefficient}) can be controlled by the extend of the bouncing mechanism (Figure \ref{fig:BLCC_by_bounce}). 

\begin{figure}[htbp]
	\centering
		\includegraphics[width=\textwidth]{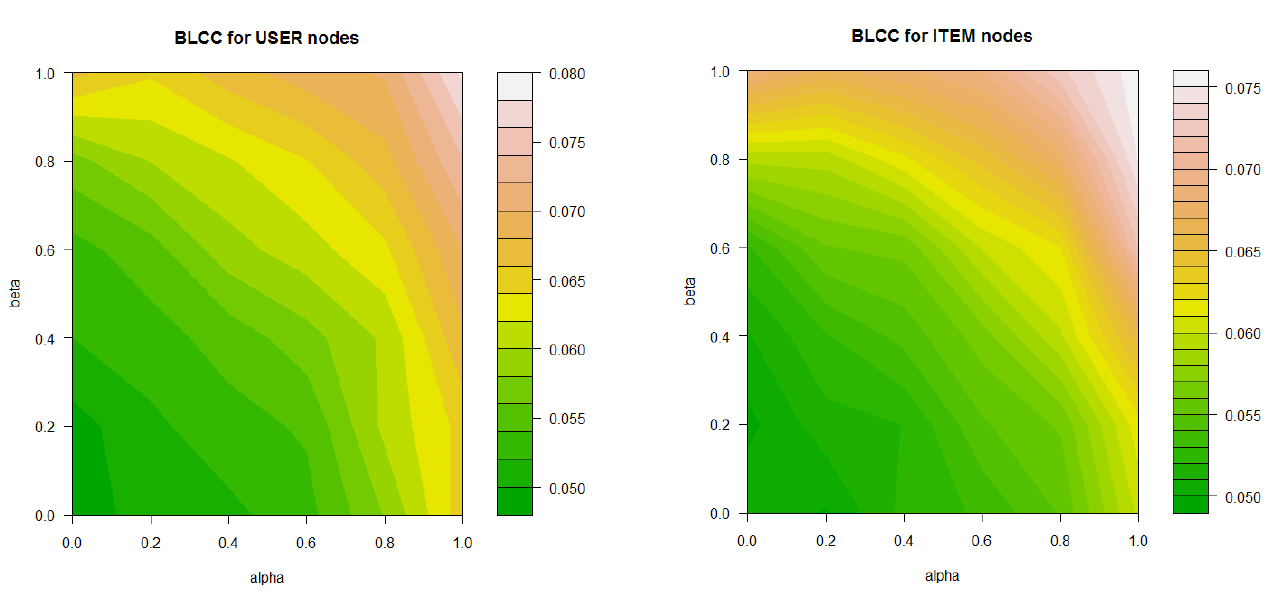}
	\caption{BLCC growths as more edges are connected with preferential attachment mechanism. The phenomenon is observed even when the bouncing parameter is zero. Experiments run with ($m=50$, $T=10~000$, $p=0.5$, $u=v=7$, $b=0$).}
	\label{fig:BLCC_by_alfa_beta}
\end{figure}

If we neglect the bouncing mechanism ($b=0$) BLCC is controlled by node degree distributions (Figure \ref{fig:BLCC_by_alfa_beta}).

There exist several other network properties that can be tunned by the parameters in our model. Such as an average distance between randomly selected pairs of nodes, the diameter of a bigraph, resilience to attack, spread of innovations or creation of the largest connected component. We omit the analysis of these features as they do not seem to have direct impact on the performance of the recommender systems.

\subsection{Neighborhood size properties}

\begin{figure}[htbp]
\begin{tabular}{cc}
	\centering
	\includegraphics[width=0.5\textwidth]{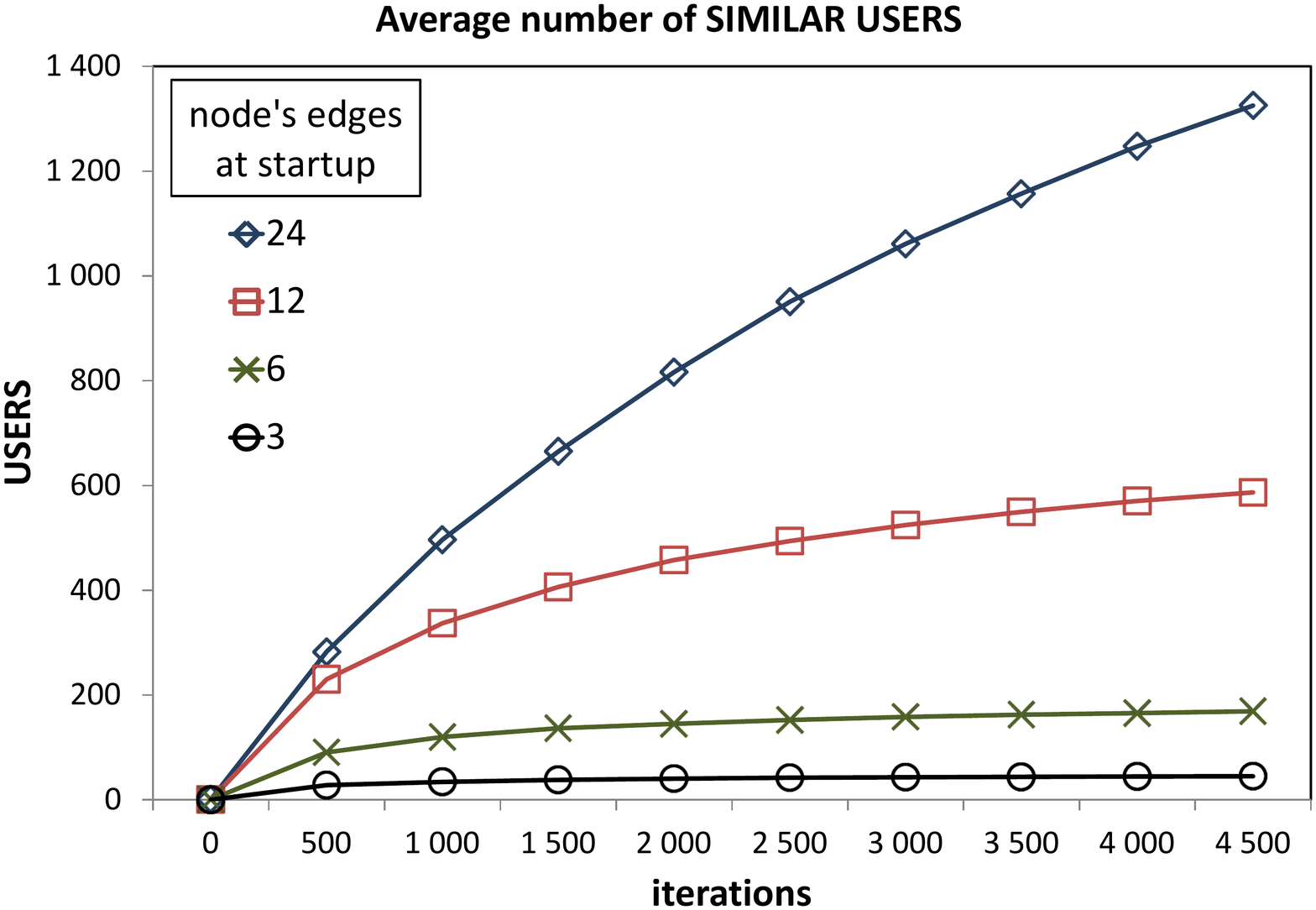}&\includegraphics[width=0.5\textwidth]{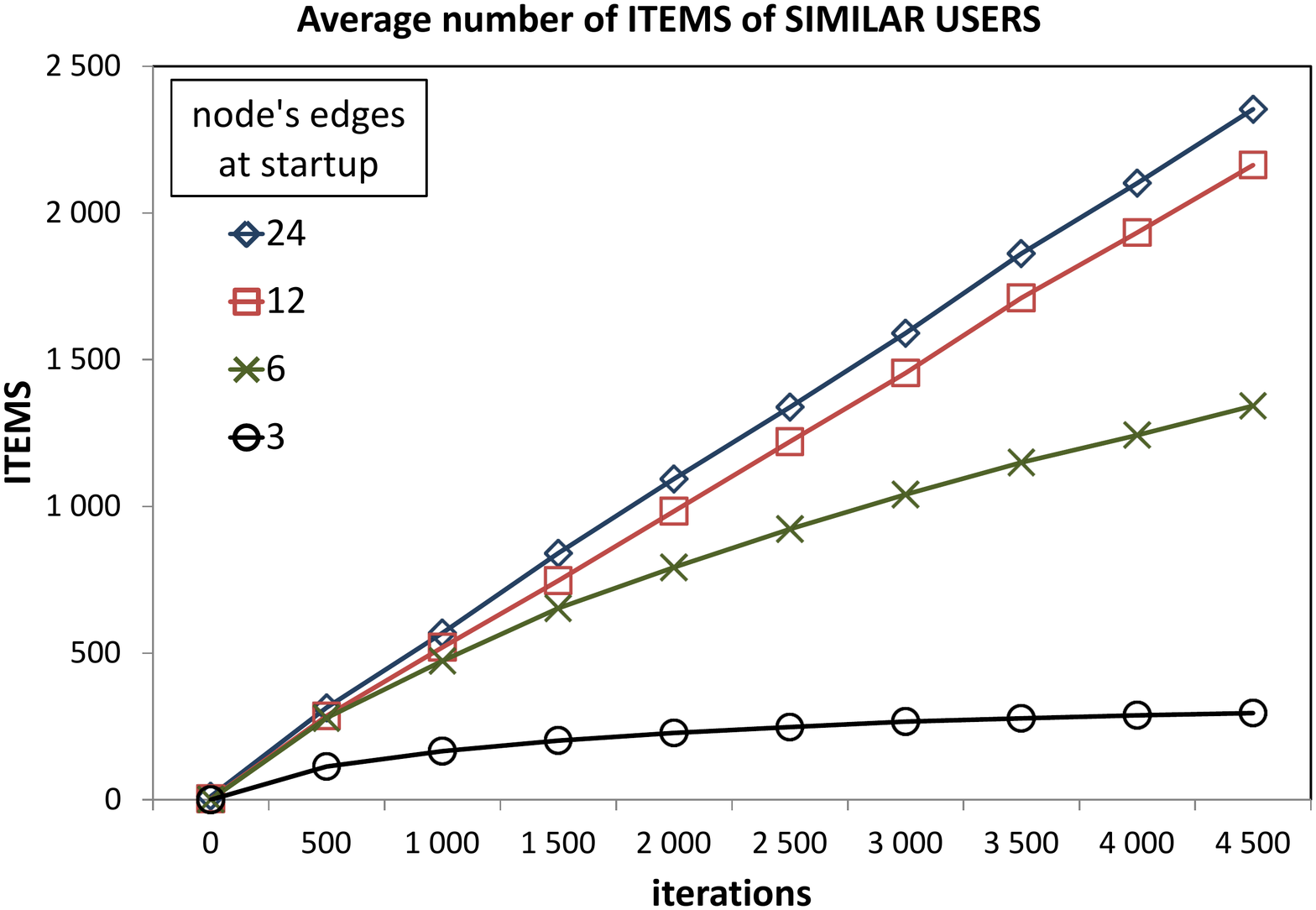}\\
	\end{tabular}	\caption{An average number of similar users (having at least one common item with a considered user) follows the growth in a graph's size. The positive relation is stronger in case of the number of the items of the similar users. The density of a graph (modeled by the number of startup edges) has even stronger impact on the size of the neighborhood than the size of a graph. Experiments run with ($m=50$, $T=10~000$, $p=0.5$, $\alpha=\beta=0.5$, $b=0$).}
	\label{fig:iterations}
\end{figure}

The number of operations that a neighborhood recommender system has to perform is related to the number of similar users and the number of their items. We recommend a new item to analyzed user from the items of the users that are similar to her/him. In Figure \ref{fig:iterations} we show two intuitive results:
\begin{itemize}
	\item the size of the neighborhood grows with the size of a graph
	\item the size of the neighborhood grows with the density of a graph (fixed number of nodes and growing number of edges)
\end{itemize}

The growth of the neighborhood is relatively sharper in case of the number of items. It is interesting that the number of similar users becomes stable earlier for sparser graphs ($3$ and $6$ edges at startup) than for denser graphs ($12$ and $24$ edges at startup). 

\begin{figure}[htbp]
	\centering
		\includegraphics[width=\textwidth]{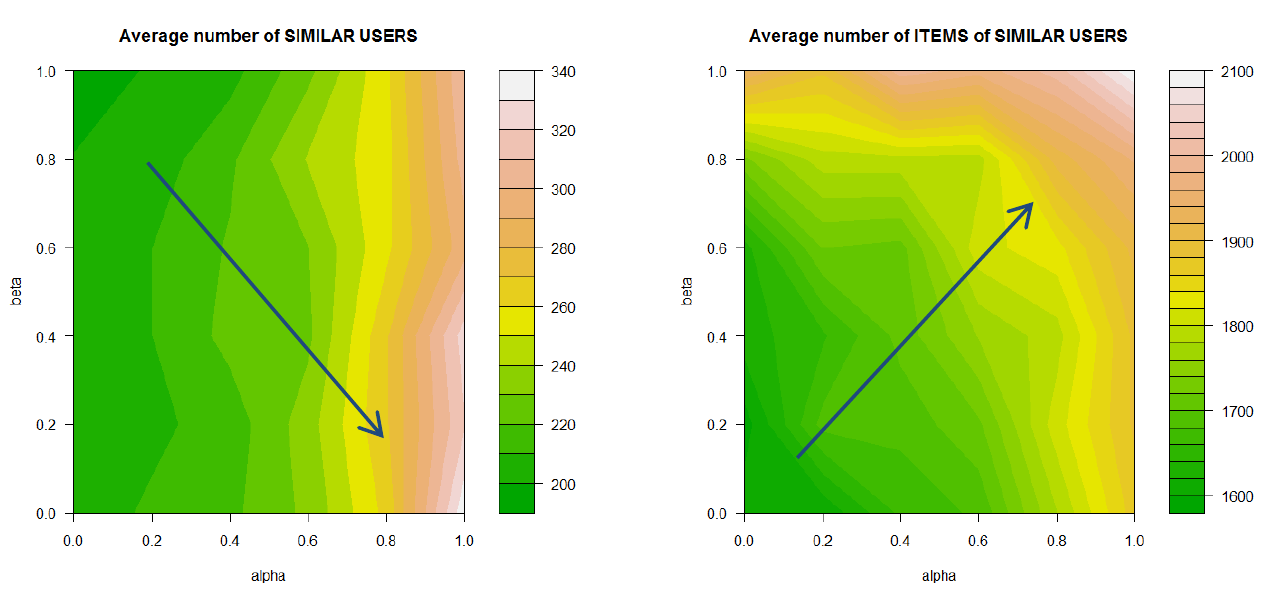}
	\caption{The shape of node degree distributions of both modalities has opposite influence on the average number of similar users. The more power-law like item degree distribution, the more neighbors can be observed. The more heavy-tailed the distribution of user nodes the stronger shrinking of the neighborhood is obtained. The arrows indicate the direction of growth. Experiments run with ($m=50$, $T=10~000$, $p=0.5$, $u=v=7$, $b=0$).}
	\label{fig:similar_by_alfa_beta}
\end{figure}

A result of potentially great importance is drawn in Figure \ref{fig:similar_by_alfa_beta}. It turns out that the impact of the shapes of node degree distributions (controlled by parameters $\alpha$ and $\beta$) on the sizes of the neighborhoods is not monotonic. It turns out that the more exponential like than power-law like the distribution of users' degrees the smaller number of similar users is observed. In all other cases the opposite force is identified.

\begin{figure}[htbp]
\begin{tabular}{cc}
	\centering
	\includegraphics[width=0.5\textwidth]{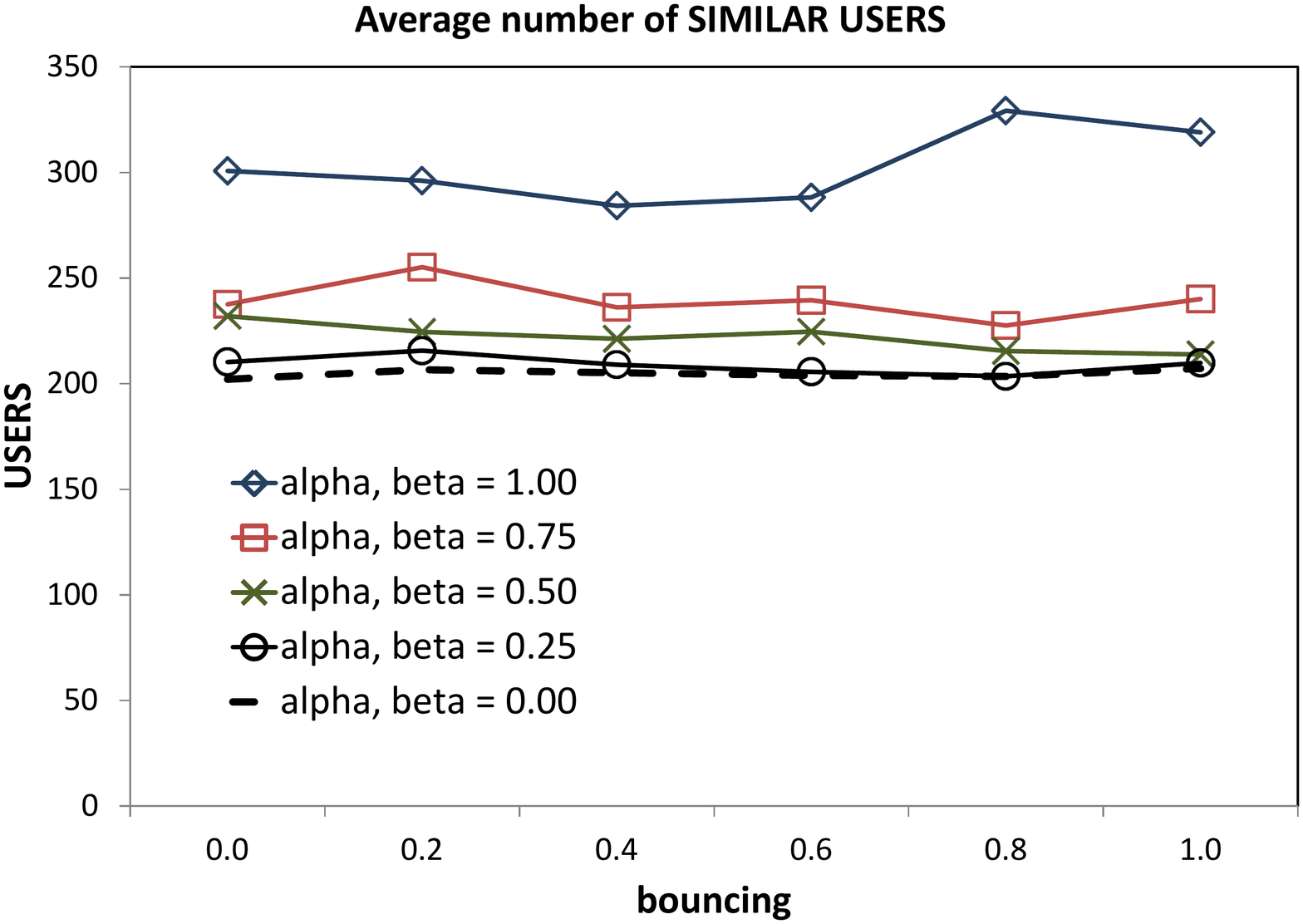}&\includegraphics[width=0.5\textwidth]{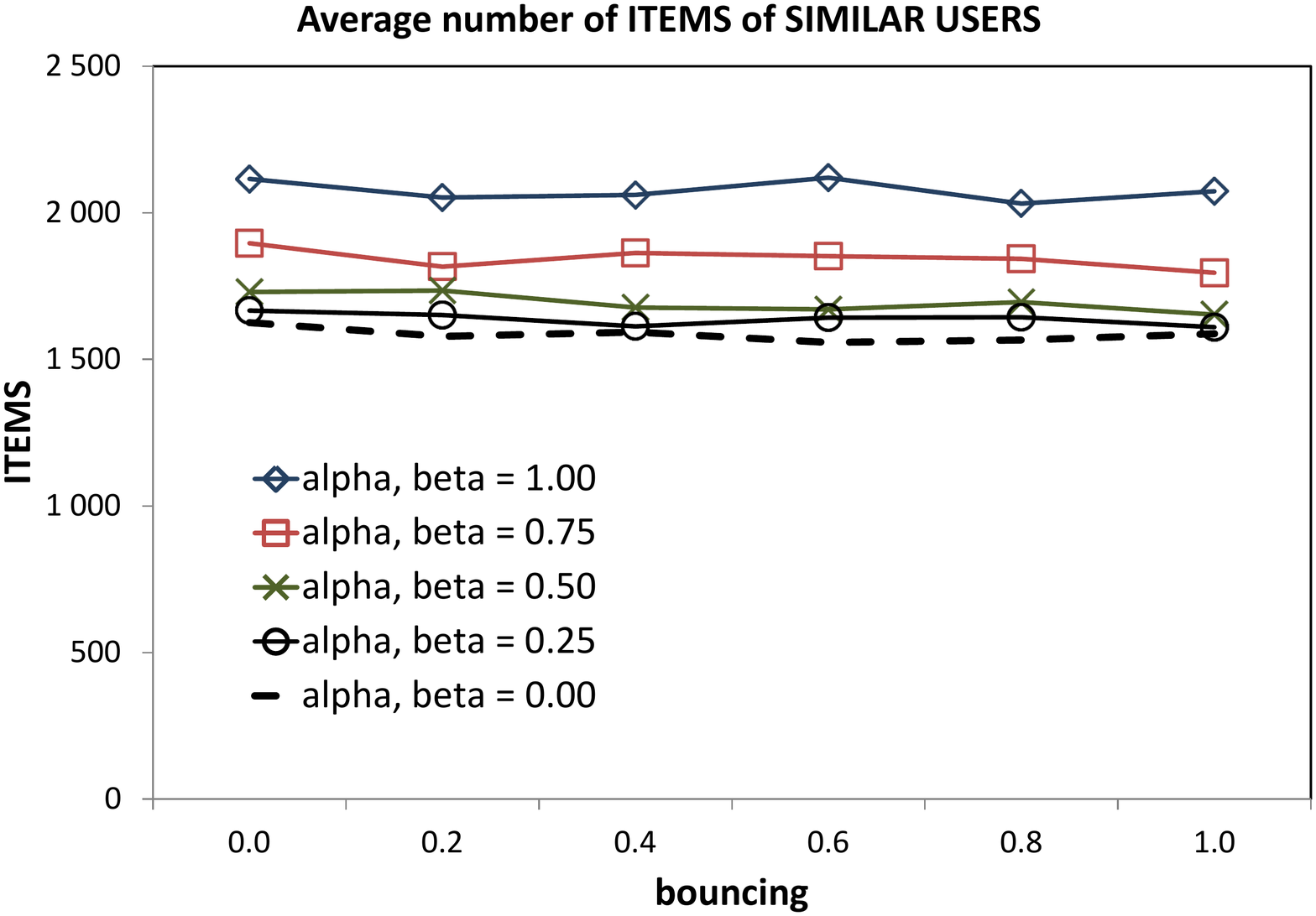}\\
	\end{tabular}		\caption{The growth of the bouncing parameter $b$ has slight negative impact of the size of both neighborhoods. However, the number of similar users and their items is determined mostly by the shapes of node degree distributions.}
	\label{fig:similar_bounce}
\end{figure}

The result presented in Figure \ref{fig:similar_bounce} is somewhat disappointing. The shrinking impact of the bouncing mechanism on the sizes of the neighborhoods is hardly observed. The effect of bouncing is too gentle compared to the level at which we are placed by the power-law distribution. Also random changes among various networks are stronger at the level than the shrinking forces. This drawback reflects the fact that in growing random graphs positive clustering coefficient is correlated with power-law node degree distribution and we are unable to generate graphs with both the exponential node degree distribution and high value of the clustering.

\section{Conclusion}

We have presented a new random graph generative algorithm dedicated to modeling performance of recommender systems. We have shown that the parameters of the algorithms influence not only pure network properties of created bigraphs, but also the properties related to the performance of neighborhood based collaborative filtering systems. Besides of the above features, the procedure enables us to output bigraphs of different sizes, densities and the proportions of the number of users to the number of items. We plan to compare how various features of bigraphs impinge on time and memory requirements of existing systems. Consequently, better understand the algorithms, their implementations and finally improve both of them.

\newpage
\subsubsection{Acknowledgments.}
This work was partially supported by Polish state budget funds for scientific research within research project \textit{Analysis and visualization of structure and dynamics of social networks using nature inspired methods}, grant No.  N516 443038.

\appendix
\section{Degree of a neighboring node}
\begin{figure}[htbp]
	\centering
		\includegraphics[width=0.6\textwidth]{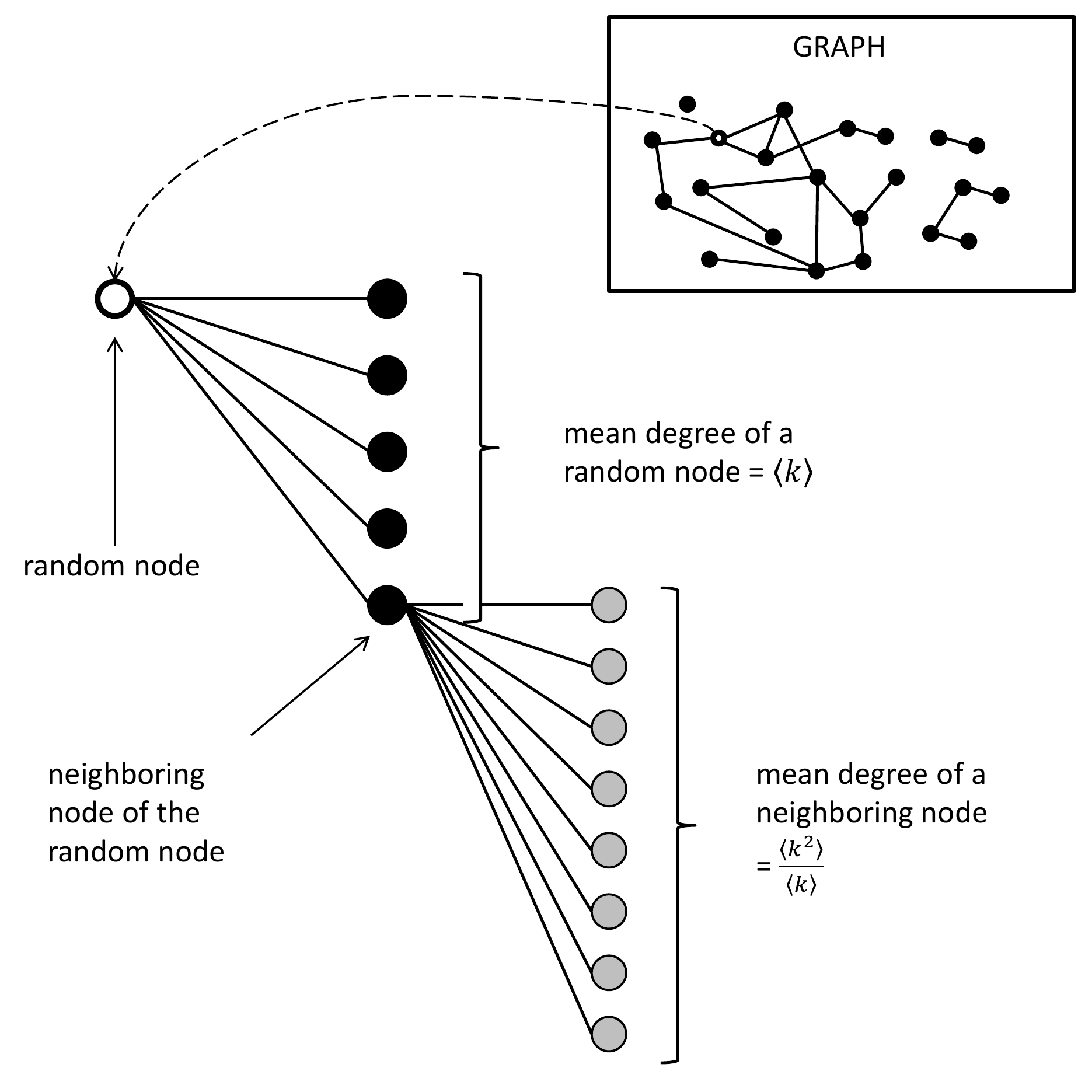}
	\caption{The expected degree of a neighbor of randomly selected node is larger than an average node degree.}
	\label{fig:neighboringNode}
\end{figure}

In this appendix we derive the expected degree of a \textit{neighboring node} in a random graph (Figure \ref{fig:neighboringNode}). Let's denote by $\langle k \rangle$ and $\langle k^2 \rangle$ the first and the second moments of the node degree distribution of graph $G=(V,E)$.

If we pick a random node from a graph then its expected number of neighbors (degree) is $\langle k \rangle$. Each of $\langle k \rangle$ edges points at a different vertex. The probability that a random edge is connected to a node is proportional to the total number of edges that are connected with the node. The probability that a random edge is connected to a node $i$ with degree $k_i$ is equal to $\frac{k_i}{\sum_{j\in V}{k_j}}$. Hence, the expected degree of a neighboring node is:

\begin{equation}
\sum_{i\in V}{k_i\frac{k_i}{\sum_{j\in V}{k_j}}}=\frac{\sum_{i\in V}{k_i^2}}{\sum_{j\in V}{k_j}}=\frac{\langle k^2 \rangle}{\langle k \rangle}.
\label{neighbor}
\end{equation}
The analysis is based on an assumption that there exist no correlation between the degrees of two neighboring nodes. 

We can show that this value is not smaller than $\langle k \rangle$ i.e. an expected degree of a random node.
Let us recall the Cauchy-Schwartz inequality:
\begin{equation}
\left(\sum_{i=1}^n x_i y_i\right)^2\leq \left(\sum_{i=1}^n x_i^2\right) \left(\sum_{i=1}^n y_i^2\right).
\label{neighbor}
\end{equation}

By putting $x_i=1$ for $i=1, \dots ,n$, we get:  

\begin{equation}
\left(\sum_{i=1}^n y_i\right)^2\leq n \left(\sum_{i=1}^n y_i^2\right),
\label{neighbor}
\end{equation}

\noindent and

\begin{equation}
\frac{\sum_{i=1}^n y_i}{n}\leq \frac{\left(\sum_{i=1}^n y_i^2\right)/n}{\left(\sum_{i=1}^n y_i\right)/n} \Rightarrow \langle y \rangle\leq\frac{\langle y^2 \rangle}{\langle y \rangle}.
\label{neighbor}
\end{equation}

\section{Node degree distribution}
We follow \textit{continuum approach} \cite{bar99a} to derive user node degree distribution. The item node degree distribution can be obtained analogously. The calculations consist of three steps. Firstly, let's solve Eq. (\ref{basicDif}).  

\begin{align*}
\frac{\partial k_j}{\partial t}&=(1-p)v\Pi(k_j)\\
&=(1-p) v  \left( \frac{\beta}{pt}+\frac{(1-\beta)k_j}{\eta t}\right)\\
&=(1-p) v  \frac{1}{t}\left( \frac{\beta  \eta  + p  (1-\beta)k_j}{p  \eta }\right),
\end{align*}

\noindent which yields

\begin{equation}
\int{\frac{1}{(1-p) v} \cdot \frac{p \eta }{\beta \eta  + p(1-\beta)k_j}}\, dk_j =\int{ \frac{1}{t}}\, dt.
\label{step2}
\end{equation}

Taking into account an initial condition $k_j(t_j)=u$, where $t_j$ is the time of creating user $j$, and the fact that $\int \frac{c}{ax+b}dx=\frac{c}{a}\ln|ax+b|+C$ we obtain

\begin{equation}
 \frac{p \eta }{(1-p) v p (1-\beta)} \left( \left[\ln \left(\beta \eta  + p(1-\beta)k_j \right) \right]
 - \left[\ln \left(\beta \eta  + p(1-\beta)u \right) \right] \right)= \left[\ln t \right] - \left[\ln t_j \right],  
\end{equation}

\noindent both sides of which can be used as exponents of $e$, giving

\begin{equation}
 \left( \frac{\beta \eta  + p(1-\beta)k_j}{\beta \eta  + p(1-\beta)u} \right) ^{\frac{\eta }{(1-p)(1-\beta)v}}
 = \left( \frac{t}{t_j} \right),
\end{equation}

\noindent after reorganizing, we have

\begin{equation}
 k_j(t)=\frac{1}{p(1-\beta)} \cdot \left( \left( \beta \eta  +p(1-\beta)u \right)\left( \frac{t}{t_j}\right)^{\frac{(1-p)(1-\beta)i}{\eta }}-\beta \eta  \right).
\end{equation}

The probability that $k_j$ is smaller then a given $k$ is:

\begin{equation}
\Phi \left\{  k_j(t)<k \right\}=\Phi \left\{\frac{\left( \beta \eta  +p(1-\beta)u \right)\left( \frac{t}{t_j}\right)^{\frac{(1-p)(1-\beta)v}{\eta }} -\beta \eta  }{p(1-\beta)} < k \right\},
\label{distr}
\end{equation}

\noindent and after reorganizing

\begin{equation}
\Phi \left\{ k_j(t)<k \right\}=\Phi \left\{ t_j > t \left( \frac{\beta \eta  +p(1-\beta)k}{\beta \eta  + p(1-\beta)u} \right)^{\frac{-\eta }{(1-p)(1-\beta)v}}   \right\}.
\label{distr}
\end{equation}

We can assume that nodes are added at equal time intervals until the current iteration $t$. The probability the iteration of adding node $j$ is larger than some $K \leq t$ equals $1-\Phi(t_j \leq K)=1-K\frac{1}{t}$. Substituting this assumption into Eq. (\ref{distr}), we obtain

\begin{align*}
\Phi \left\{  k_j(t)<k \right\}&=1- \Phi \left\{ t_j \leq t \left( \frac{\beta \eta  +p(1-\beta)k}{\beta \eta  + p(1-\beta)u} \right)^{\frac{-\eta }{(1-p)(1-\beta)v}}   \right\} \\
&=1-\left( \frac{\beta \eta  +p(1-\beta)k}{\beta \eta  + p(1-\beta)u} \right)^{\frac{-\eta }{(1-p)(1-\beta)v}}.\\
\label{distr2}
\end{align*}

We can obtain probability density function of random variable $k$ by differentiating its cumulative distribution function $P(k)=\partial  \Phi\{k_j(t)<k\}/\partial k$, as a result we have

\begin{equation}
P(k)=\frac{\eta }{(1-p)(1-\beta)v} \cdot p(1-\beta) \cdot \left( \frac{\beta \eta  +p(1-\beta)k}{\beta \eta  + p(1-\beta)u} \right)^{\frac{-\eta }{(1-p)(1-\beta)v}-1},
\label{distr3}
\end{equation}

\noindent that is:

\begin{equation}
P(k)\propto \left( \frac{\beta \eta  +p(1-\beta)k}{\beta \eta  + p(1-\beta)u} \right)^{\frac{-\eta }{(1-p)(1-\beta)v}-1}.
\label{distr4}
\end{equation}
\newpage
\bibliography{preparation}
\end{document}